\documentclass[letterpaper, 10 pt, conference]{ieeeconf}  
\IEEEoverridecommandlockouts                              

\usepackage[utf8]{inputenc} 
\usepackage[T1]{fontenc} 
\usepackage{cite} 
\usepackage{graphics} 
\usepackage{times} 
\usepackage{amsmath} 
\usepackage{amssymb} 
\usepackage{amsfonts}
\usepackage{mathtools} 
\usepackage{graphicx}
\usepackage{textcomp}
\usepackage{adjustbox}
\usepackage{tikz}
\usepackage{siunitx} 
\usepackage{array}
\newcolumntype{P}{>{$}r<{$}@{\,\ensuremath{\pm}\,}>{$}l<{$}}
\usepackage{wrapfig}
\usepackage{tabularx}
\usepackage{booktabs} 
\usepackage{makecell}
\usepackage{multirow} 
\usepackage{hhline}

\usepackage{tikz}
\usetikzlibrary{backgrounds, fit, shapes.geometric, positioning}
\usepackage{url}
\usepackage{xcolor}
\usepackage{soul}
\usepackage{rotating}

\usepackage{dsfont}
\usepackage{flushend}

\usepackage{pifont}

\usepackage[font=footnotesize]{caption} 
\usepackage[font=footnotesize, labelfont=footnotesize, textfont=footnotesize, labelformat=brace]{subcaption} 
\makeatletter 
\renewcommand\p@subfigure{\thefigure\,}
\makeatother 

\usepackage[hidelinks,bookmarksnumbered,bookmarksdepth=1]{hyperref} 

\usepackage[ruled, linesnumbered, vlined, noend]{algorithm2e}

\SetCommentSty{mycommfont}
\usepackage[noend]{algpseudocode}

\def\secref#1{Sec.~\ref{#1}}
\def\figref#1{Fig.~\ref{#1}}

\def\tabref#1{Tab.~\ref{#1}}

\def\eqref#1{Eq.~(\ref{#1})}

\algnewcommand{\algorithmicgoto}{\textbf{go to}}
\algnewcommand{\Goto}[1]{\algorithmicgoto~line~\ref{#1}}
\usepackage{refcount} 
\usepackage{xstring}  

\newcommand\etal{~\emph{et al.}}

\newcolumntype{Y}{>{\centering\arraybackslash}X}
\def\argmax{\mathop{\rm argmax}}

\title{\LARGE \bf OMAR: Open-World Multi-Attribute Reasoning\\for Scene-Conditioned Shelf Organization}

\author{Benedikt~Kreis$^{\text{\textasteriskcentered,1,2,3}}$ \and Rohit~Menon$^{\text{\textasteriskcentered,1,2,3}}$ \and Benno~Wingender$^{\text{1,2,3}}$ \and M.\,E.~Ullah$^{\text{1}}$ \and Nils~Dengler$^{\text{4}}$ \and Maren~Bennewitz$^{\text{1,2,3}}$
\thanks{
  	\hspace{-1.05em}$^{\text{1}}$Humanoid Robots Lab, University of Bonn, Germany.\newline
  	$^{\text{2}}$Center for Robotics, University of Bonn, Germany.\newline
  	$^{\text{3}}$Lamarr Institute for ML and AI, Bonn, Germany.\newline
    $^{\text{4}}$Interactive Robot Perception and Learning Lab, Technical University of Darmstadt, Germany.\newline
    $^{\text{\textasteriskcentered}}$These authors contributed equally to this work.\newline
    This work has been partially funded by the German Federal Ministry of Research, Technology and Space (BMFTR) under the Robotics Institute Germany, grant No. 16ME0999.
	Generative AI was used for linguistic refinement.
    The corresponding author is \mbox{Benedikt Kreis: \href{mailto:kreis@cs.uni-bonn.de}{kreis@cs.uni-bonn.de}}}%
}
\begin{document}
\maketitle

\begin{abstract}
With the advent of foundation models, robots are increasingly capable of understanding and executing explicitly specified human tasks. 
Extending these capabilities to open-ended tasks, however, requires robots to infer appropriate actions from the observed scene.
Household organization exemplifies this challenge, as robots must often act without detailed instructions, predefined object vocabularies, or known object-receptacle associations.
While robots may follow instructions or learn organization preferences from demonstrations, inferring the latent organization of an occasionally changing scene and placing a novel object consistently with it remains challenging. 
We propose an open-world multi-attribute reasoning pipeline that grounds unknown objects across views and constructs an instance-level 3D scene graph for placement reasoning. 
First, a vision-language model proposes labels together with attributes such as content and packaging type, which are grounded through segmentation, depth, and multi-view association.
Then, an LLM reasons over this representation to infer the latent shelf organization and rank scene-consistent object placements without explicitly user-provided preferences.
The resulting placement rankings guide subsequent geometric and kinematic feasibility checks for execution. 
Our experimental evaluation shows that our approach leads to improved instance mapping and placement coherence over single-view language-grounding and geometry-only baselines, enabling autonomous placement of previously unseen objects without predefined object-receptacle mappings or class-specific retraining.
\end{abstract}

\section{Introduction}
\label{sec:intro}
Households are constantly changing environments, and their organization reflects implicit and evolving human conventions.
To effectively organize objects in such environments, service robots must act beyond explicitly specified tasks.
When storing groceries, a robot must determine where a previously unseen object belongs, even without detailed instructions or user-provided preferences.
As object locations change, predefined object vocabularies and object-receptacle mappings cannot fully specify appropriate placements.

We study scene-conditioned shelf organization, in which the robot must infer the latent organization from the observed arrangement and place a novel object consistently into it.
This task requires reasoning over multiple object attributes, including content, packaging type, edibility, and orientation.
These attributes can suggest competing placements, and their relative importance depends on the organization of the particular scene.
Category labels alone therefore provide insufficient context.
Placement reasoning requires object-level descriptions grounded in the observed shelf.
Multi-view grounding provides this object-level context by associating instances across views and revealing occluded objects.

\begin{figure}[t]
    \centering
    \vspace{0.25\baselineskip}
    \includegraphics[width=0.89\linewidth, trim=215 55 220 55, clip]{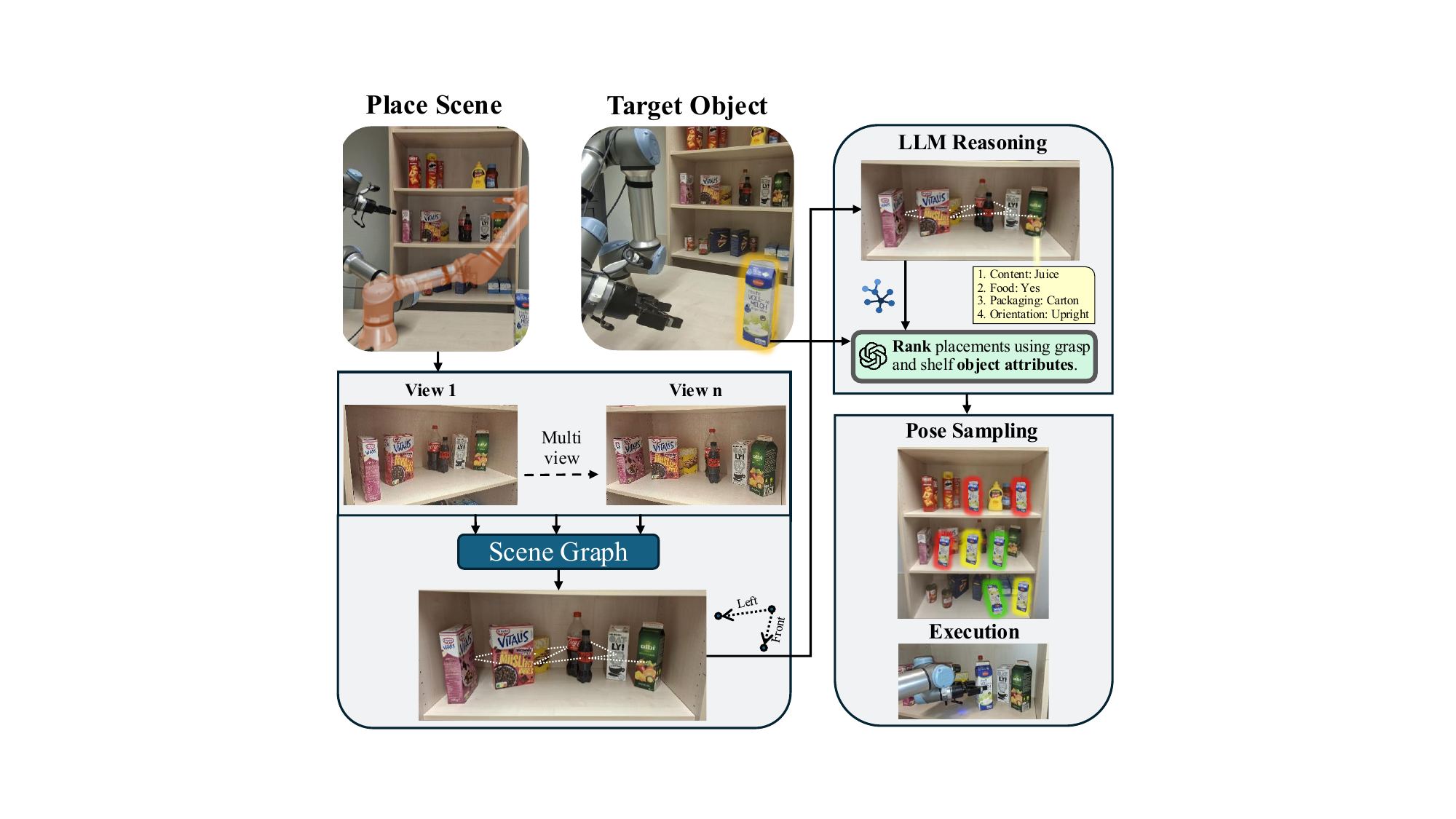}
    \vspace{-0.4\baselineskip}
    \caption{
    The robot's task is to pick an unknown object from the table and find the semantically most coherent and geometrically feasible placement in the shelving unit across multiple shelf boards (top left).
    Our system creates a 3D representation from multiple views (bottom left) to predict placement poses considering multiple object attributes (right).
    Our approach combines open-world multi-attribute and multi-view instance mapping to generate a 3D semantic scene graph, which enables scene-conditioned placement reasoning with feasibility-aware grasp-placement planning to execute placements of previously unseen objects into space-limited shelving units.
    }
    \label{fig:motivation}
\end{figure}

Vision-language models (VLMs) and large language models (LLMs) offer semantic knowledge beyond predefined object sets, supporting the prediction of semantically plausible placements~\cite{argenzianoEMPOWEREmbodiedMultirole2024,abdelreheemPlaceIt3DLanguageGuidedObject2025,xuSetItFunctional2024}, but plausibility alone does not ensure geometric or kinematic feasibility.
Joint pick-and-place pipelines address grasp-placement compatibility through task-aware 6-DoF grasp selection and feasibility evaluation~\cite{hePick2PlaceTaskAware6DoF2023b, shanthiPickPlacePlanning2024, wingenderRobustPlaceabilityMetric2026}, but do not infer the shelf's latent organization from semantic object attributes.
The resulting challenge is to connect grounded, scene-dependent semantic reasoning with feasibility-aware planning so that placements of unseen objects preserve the observed organization and remain executable in confined shelf spaces.

We address this gap via open-world, multi-attribute reasoning (OMAR), which places previously unseen objects consistently with the observed shelf organization.
As shown in \figref{fig:motivation}, our pipeline grounds VLM-generated object descriptions across views in an instance-level 3D scene graph, capturing content, packaging type, edibility, and orientation rather than category labels alone.
An LLM reasons jointly over these attributes and their scene-dependent relevance to infer the latent shelf organization and propose placements that preserve its coherence, without user-provided preferences, predefined object-receptacle mappings, or class-specific retraining.
These proposals guide grasp prediction and geometric and kinematic feasibility checks, linking multi-attribute shelf coherence to executable placements.
To summarize, the main contributions of our work are:

\begin{itemize}
    \item An open-world multi-attribute mapping pipeline that grounds unknown objects across views and creates an instance-level 3D scene graph for placement reasoning.
    \item An LLM-based reasoning pipeline leveraging instance-level scene representation to infer the latent shelf organization and rank scene-consistent object placements.
\end{itemize}

Our evaluation shows that our approach leads to improved instance mapping and placement coherence over single-view language-grounding and geometry-only baselines.

\section{Related Work}
\label{sec:related}
Picking and placing (P\&P) previously unseen household objects is a challenge at the intersection of perception and manipulation.
Recent advances in transformer-based foundation models~\cite{sanghaiAdvancesTransformersRobotic2024} enable broader generalization to unseen tasks and objects compared to classical approaches.

\subsection{Classical and Learning-based P\&P Approaches}
\label{subsec:classical_approaches}
Pick2Place~\cite{hePick2PlaceTaskAware6DoF2023b} uses object-centric perspective affordances to produce task-aware 6-DoF grasps that align grasp selection with downstream placements.
Alternatively, cost-based joint pick-and-place formulations produce place-aware grasps, as demonstrated by Shanthi and Hermans~\cite{shanthiPickPlacePlanning2024} and, more recently, UniP~\cite{wingenderRobustPlaceabilityMetric2026}, which introduced a generalized placeability metric that scores object configurations based on geometric and physical stability criteria.
With a similar emphasis on physical stability, Paxton\etal~\cite{paxtonPredictingStableConfigurations2022} learn stable object placements.
Moving from individual placements to multi-step rearrangement, REAP~\cite{swobodaREAPLearningGeneral2026} uses a heuristic learned from motion-planner feedback, reducing motion-planning queries while guiding search toward geometrically feasible actions.

Beyond physical feasibility, semantic placement planning is considered by Wingender\etal~\cite{wingenderSemanticDensityAwarePlanning2026}, although the approach uses predefined semantic labels.
Similarly, Adeleye\etal~\cite{adeleyePuttingAwayGroceries2022} combine object recognition with semantic reasoning to select grocery storage locations in shelf environments, but operate on a closed object set without checking geometric feasibility.
These approaches thus either omit placement semantics~\cite{paxtonPredictingStableConfigurations2022, hePick2PlaceTaskAware6DoF2023b, shanthiPickPlacePlanning2024, wingenderRobustPlaceabilityMetric2026} or rely on predefined object sets without considering multiple object attributes~\cite{wingenderSemanticDensityAwarePlanning2026, adeleyePuttingAwayGroceries2022}.
Furthermore, some approaches~\cite{shanthiPickPlacePlanning2024, paxtonPredictingStableConfigurations2022} do not address placements into geometrically challenging confined spaces.

\subsection{Foundation Model-based P\&P Approaches}
\label{subsec:llm_approaches}
The generalization promise and semantic reasoning capabilities drives the increased use of foundation models in robotic manipulation.
For example, TidyBot~\cite{wuTidyBotPersonalizedRobot2023} combines open-vocabulary detection using ViLD~\cite{guOpenvocabularyObjectDetection2021} with object-centric manipulation, using sparse user preferences to bootstrap personalized object rearrangements.
ContextSortLM~\cite{ramachandruniPersonalizedRoboticObject2025} and APRICOT~\cite{wangAPRICOTActivePreference2024a} similarly use language models to learn detailed user-specific spatial preferences for object placements.
While these approaches demonstrate the value of foundation models for semantic placement decisions, their reasoning requires user specific input.

ConceptGraphs~\cite{guConceptGraphsOpenVocabulary3D2024} and BBQ~\cite{linokBareQueriesOpenVocabulary2025} use 3D scene graphs as LLM input for language-guided object selection and relational object grounding, respectively.
These works ground language-model reasoning in observed objects and their spatial context through scene-graph representations.
Rather than grounding a user-specified object query, we reason over scene-graph attributes to infer shelf organization and rank placements of unseen objects.

Foundation models have also been applied specifically to task-oriented grasping.
GraspGPT~\cite{tangGraspGPTLeveragingSemantic2023} and FoundationGrasp~\cite{tangFoundationGraspGeneralizableTaskOriented2025} leverage semantic knowledge to bias grasp selection towards task-relevant affordances, while GraspMolmo~\cite{deshpandeGraspMolmoGeneralizableTaskOriented2025} combines a multimodal foundation model with large-scale synthetic data to improve generalization.
These methods primarily reason about how an object should be grasped, rather than how it should be placed.

Conversely, several recent approaches focus on semantic placement.
AnyPlace~\cite{zhaoAnyPlaceLearningGeneralizable2025} learns generalizable object-to-receptacle placements from multimodal supervision, PlaceIt3D~\cite{abdelreheemPlaceIt3DLanguageGuidedObject2025} reasons about language-guided placements in real 3D scenes but without robotic embodiment, and Set It Up~\cite{xuSetItFunctional2024} targets functional object arrangements.
These methods produce semantically meaningful placements but only partially address geometric and kinematic constraints.

Integrated foundation-model-based systems that focus on confined spaces are LLM-Pack~\cite{bleiLLMPackIntuitiveGrocery2025} which uses VLMs to identify grocery items and an LLM to generate human-like packing sequences.
EMPOWER~\cite{argenzianoEMPOWEREmbodiedMultirole2024} employs LLM agents for perception, planning, and execution, enabling multi-step object rearrangements through a set of predefined manipulation skills.
However, these systems do not explicitly couple semantic placement reasoning with the grasp-dependent geometric feasibility of executing individual placements in space-limited environments.

Our approach addresses the existing gaps by coupling open-world, foundation-model-based semantic reasoning with grasp-placement compatibility.
A persistent multi-view instance map provides a holistic and instance-consistent representation of complex scenes across viewpoints and enriches objects with detailed semantic attributes beyond category labels, enabling more informed placement decisions.
This allows to jointly consider semantic and geometric fit, reachability, and grasp feasibility across multiple shelf boards with varying clearance.
\begin{figure*}[t]
    \centering
    \includegraphics[width=0.96\linewidth]{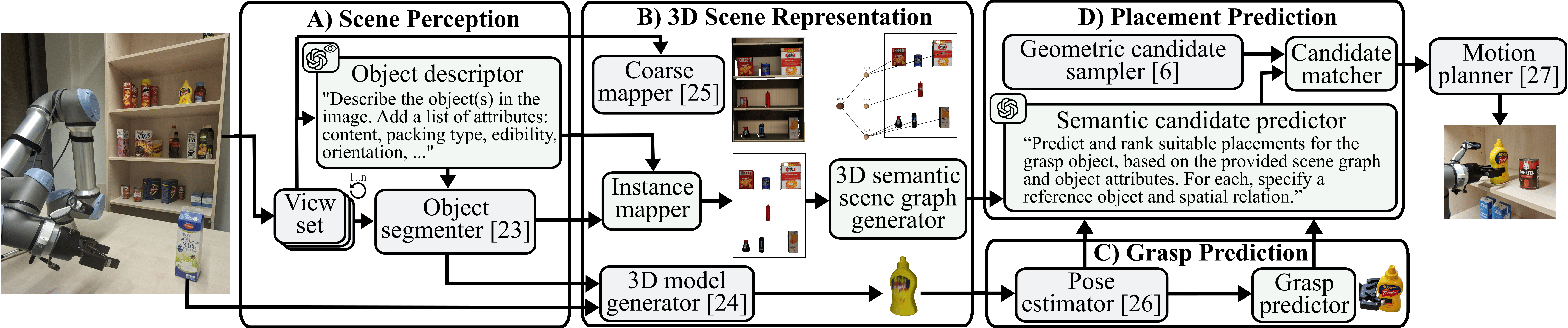}
    \vspace{-0.25\baselineskip}
    \caption{
    Overview of the proposed approach:
    First the robot discovers the environment (A) by moving the camera to identify the unknown grasp object presented on the table and then to the shelf to create an object instance map (B) using descriptions and masks of each object from the view set.
    Based on the instance map, we create a 3D scene graph, containing semantic and geometric attributes of each object.
    Furthermore, we create a 3D model of the target object, to estimate its pose and evaluate valid grasps (C).
    Afterwards, we predict placement candidates across all shelf boards (D) by sampling geometrically possible candidates on the map and match them with semantically suitable candidates predicted based on the 3D semantic scene graph.
    The matched candidates are ranked and based on multiple feasibility checks unsuitable candidates are discarded.
    Finally, a motion planner calculates collision-free trajectories for the remaining candidates (grasp-placement pairs) and executes the highest-scored one.
    }
    \label{fig:architecture}
\end{figure*}

\section{Task Description and Assumptions}
\label{sec:problem_assumptions}
We consider the following problem: Given a shelf scene containing an open-set of packaged grocery and household objects on multiple shelf boards, a robotic arm must pick an unknown target object from a table and place it at the semantically most coherent location on the shelving unit while ensuring physical feasibility.
We restrict placement orientations by packaging type to prevent objects such as bottles and cans from rolling, permitting lying poses only for packaging with planar side faces and assuming stable support in the allowed orientations.
During planning and execution, we assume static scenes observable with an \mbox{RGB-D} camera mounted on the robot arm, which allows to perceive the place scene (shelf) and the target object from multiple view points.

\section{Our Approach}
\label{sec:approach}
We propose an open-world pipeline for placing a previously unseen object in shelf environments leveraging multi-view semantic instance mapping.
Converting the map into a 3D semantic scene graph representation allows LLM-based multi-attribute reasoning of placements, which we combine with target-oriented antipodal grasp generation to select grasp-placement pairs to maintain the latent semantic shelf organization.
\figref{fig:architecture} provides an overview of the pipeline, whose main components are detailed below.

\subsection{Autonomous Open-World Scene Perception}\label{subsec:exploration}
To reason about placements without a predefined object vocabulary, the robot must obtain more than category labels: it requires attributes that both distinguish instances and expose the organization of the observed shelf.
We therefore acquire RGB-D observations of the unknown target object and the shelving unit and query a VLM at each of the \(n\) viewpoints using our structured object descriptor.

For each object, the VLM predicts a textual label and multiple attributes such as content (e.g., crackers, mustard), packaging type (e.g., box, bottle), edibility (food or drink versus non-food items), and orientation (e.g., upright, lying).
Applying the same schema to the target and shelf objects creates a shared semantic representation for cross-view association and subsequent placement prediction.
Unlike label-only perception~\cite{adeleyePuttingAwayGroceries2022, argenzianoEMPOWEREmbodiedMultirole2024, wingenderSemanticDensityAwarePlanning2026}, this attribute-rich, open-world description provides complementary semantic and physical cues without class-specific retraining or predefined object-receptacle mappings.

We use the predicted labels as prompts for the SAM\,3 segmentation model~\cite{carionSAM3Segment2026} and thereby ground each description in an image mask and its corresponding depth measurements.
The resulting RGB-D observations, masks, and semantic predictions form an attribute-annotated view set that is updated as the robot moves across viewpoints.

In parallel, we generate a 3D mesh of the target object from a single RGB-D observation using the generative flow-matching model SAM\,3D~\cite{chenSAM3D3Dfy2026}.
This mesh is used later for grasp generation and geometric placement validation.
The view set, however, remains an aggregation of partial and potentially inconsistent observations.
To obtain the persistent object-level context required for placement reasoning, we next consolidate these observations across views into one unified 3D representation.

\subsection{Multi-View 3D Scene Representation}\label{subsec:mapping}
Integrating multiple viewpoints leverages close-up views to reveal finer object details, enabling, e.g., more reliable inference of object content.
Furthermore, occlusions make individual views incomplete, and viewpoint-dependent VLM descriptions may assign different labels or attributes to the same object.
We therefore retain labels proposed for the same object across views as alternative hypotheses, accumulating evidence before selecting a canonical label.
Naively fusing these observations can duplicate instances, merge distinct objects, or attach inconsistent semantics, distorting the shelf organization presented to the placement predictor.

Our scene representation address these ambiguities in two stages.
First, a multi-view semantic instance mapper combines geometric association with VLM-derived extent cues and accumulates alternative label hypotheses instead of committing to a single per-view prediction.
Second, the resulting persistent instances are converted into an instance-level 3D semantic scene graph that jointly encodes object attributes, geometry, and spatial relations.
Unlike individual images, the scene graph consolidates partial observations into a view-independent representation with explicit object identities and spatial relations, avoiding the need for repeated inference of occluded, viewpoint-dependent observations.
The two-stage association and its conversion into a multi-attribute scene graph provides the structured, scene-wide context used to infer the latent shelf organization.

\subsubsection*{\textbf{Multi-View Semantic Instance Mapping}}
We construct a 3D semantic instance map \(\mathcal{M}_{\text{inst}}\) in which each object instance \(i\) has a dedicated fine-grained TSDF volume, semantic attributes, and a set of label hypotheses \(\mathcal{H}_i\).
Maintaining these hypotheses across views separates object identity from viewpoint-dependent descriptions, allowing instance labels to be revised without fragmenting the instance map.

For each detection \(d\), candidate instances are filtered by bounding-box center distance and 3D overlap.
We measure 3D overlap by dividing the intersection volume of the detection and instance bounding boxes by the volume of the smaller box.
Thus, a partial detection fully contained within the accumulated instance box receives an overlap score of \(1\), even when the boxes differ greatly in size.
We also compare the labels and attributes of the detection and candidate instance, but a semantic disagreement rules out the match only when both semantic estimates are supported by consistent observations and sufficiently confident.
When semantic evidence is uncertain, we retain candidates that satisfy the geometric matching criteria for further association.

Let \(c_d\) and \(c_i\) denote the bounding-box centers of detection \(d\) and candidate instance \(i\), respectively.
We define the association cost as \(C(d,i)=\|c_d-c_i\|_2+C_{\mathrm{ext}}(d,i)\),
where the extent penalty \(C_{\mathrm{ext}}(d,i)\) measures disagreement between observed dimensions and the VLM-derived typical extent associated with the candidate instance.
This penalty provides a physical-size cue for distinguishing candidate instances whose centers lie at similar distances from the detection.
Greedy one-to-one assignment repeatedly selects the lowest-cost pair among unassigned detections and instances.

For each match, we integrate the masked RGB-D observation into the instance TSDF.
Let \(A_i^f(\ell)\) denote the accumulated evidence for label \(\ell\) of instance \(i\) after processing RGB-D frame \(f\).
For the label \(\ell_d\) of detection \(d\) in frame \(f\), we update its accumulated evidence as
\(A_i^f(\ell_d)=A_i^{f-1}(\ell_d)+s_d\,w_{\mathrm{ext}}(d,\ell_d)\),
where \(s_d\) is the segmentation confidence and \(w_{\mathrm{ext}}\) is a Gaussian extent-consistency weight that compares sorted observed and VLM-derived typical extents using a heuristically chosen \(\sigma_{\text{ext}}\)\,\(=\)\,\(0.04\), downweighting size-inconsistent evidence without hard thresholds to accommodate partial visibility and segmentation errors.
We then select the canonical label \(\ell_i^\star=\argmax_{\ell\in\mathcal{H}_i}A_i^f(\ell)\) and retain supported alternative hypotheses.
This allows repeated supporting observations to outweigh isolated misclassifications while preserving hypotheses for later revision.
Unmatched detections initialize new instances, while unmatched instances remain in the map for subsequent views.

We merge instances with a 3D bounding-box overlap of at least \SI{50}{\percent} of the smaller box's volume, subject to a semantic conflict check.
Requiring overlap of at least half the smaller box makes a merge less likely to result from incidental overlap between neighboring objects, while accommodating an incompletely reconstructed instance.
A conflict occurs when any attribute shared by both instances, such as content, packaging, edibility, or orientation, differs after normalization.
If none of these attributes is shared, a conflict requires both canonical labels to be nonempty and different after normalization.
To guard against isolated or low-confidence predictions, such a conflict prevents merging only when each instance's canonical label has at least two supporting observations, including one with segmentation confidence of at least \SI{70}{\percent}.
Repeated support reduces the influence of a single-view misclassification, while the confidence threshold limits semantic vetoes to relatively strong predictions.

The resulting 3D semantic instance map \(\mathcal{M}_{\text{inst}}\), with its instance geometry, object attributes, and label set \(\mathcal{L}\), provides the object context encoded in the subsequent 3D semantic scene graph for multi-attribute placement reasoning.

\subsubsection*{\textbf{3D Semantic Scene Graph Generation}}
\begin{figure}[t]
    \vspace{0.25\baselineskip}
    \centering
    \includegraphics[width=0.99\linewidth]{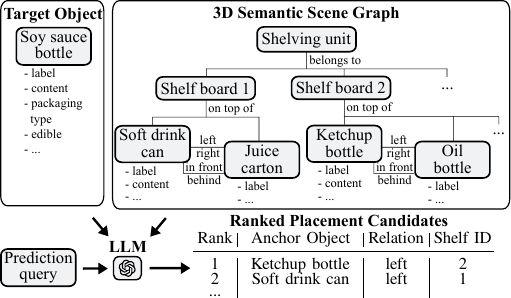}
    \vspace{-1.35\baselineskip}
    \caption{
    Based on the 3D semantic instance map of the shelving unit, we generate a 3D semantic scene graph representation (top right), which contains detailed semantic and geometrical attributes of each object instance.
    We assign the same attributes to the target object based on the RGB-D observation from the table (top left).
    Providing these inputs to the LLM, together with the prediction query (bottom left), it predicts ranked placement candidates (bottom right). 
    }
    \label{fig:scene_graph}
\end{figure}
We construct a 3D semantic scene graph from the multi-view semantic instance map, as exemplified in \figref{fig:scene_graph}.
Each object node stores semantic attributes, including its label, content, packaging type, edibility, and orientation, together with geometric attributes, including its bounding-box dimensions and centroid estimated from the instance TSDF volume.

Directed edges encode the shelf hierarchy: each shelf board \textit{belongs to} the shelving unit, and each object is linked to its assigned board by an \textit{on top of} edge, restricting comparisons to a common board.
We assume known shelf-board heights and assign each object to the board whose height is closest to its centroid height.
Also, each neighboring object pair on the same board is connected via directed edges.
Ordering centroids along the lateral axis of the board's coordinate frame determines \textit{left of}/\textit{right of} relations, while ordering by depth measured from the board's front determines \textit{in front of}/\textit{behind of} relations.
Each relation is represented together with its inverse directed edge.
For example, a \textit{left of} edge from a bottle to a box is paired with a \textit{right of} edge from the box to the bottle.
Centroid ordering provides a shelf-referenced spatial description without relying on surface-contact estimates from partially reconstructed geometry.

\subsubsection*{\textbf{Coarse Mapping}}
In addition, we fuse all RGB-D observations into a coarse TSDF map \(\mathcal{M}_\text{coarse}\)~\cite{correiamarquesOverconfidenceProblemSemantic2024} for collision-aware planning.

\subsection{Model-Aware Grasp Prediction}
\label{subsec:grasp_prediction}
We sample antipodal grasp candidates on the target object mesh using grids over its bounding-box faces, up to six approach directions, and rotations about each approach axis.
The grid sampling provides systematic geometric coverage, while perturbations of up to \SI{\pm 3}{mm} along the gripper approach, closing, and binormal axes add grasp candidates between grid samples.

To evaluate these candidates in the scene, FoundationPose~\cite{wenFoundationPoseUnified6D2024a} estimates the object's 6-DoF pose to transform candidates into the world frame, where collision and inverse-kinematics checks reject obstructed or unreachable grasps.
Then we rank the remaining candidates using a geometric quality score \(G_{\mathrm{qual}}\) to favor antipodal finger-to-object contact-normal alignment, finger-pad centering, and an object enclosing aperture.
A vertical approach preference biases selection towards tabletop grasps.

\subsection{Multi-Attribute-Aware Placement Prediction}
\label{subsec:placement_prediction}
To generate shelf placements that are coherent with the observed scene across multiple attributes, we couple LLM-based reasoning with mesh-based placement sampling.
As illustrated in \figref{fig:scene_graph}, the LLM receives a prediction query, the 3D semantic scene graph with observed object attributes, such as content, packaging type, edibility, and orientation, and the corresponding target-object attributes (see \secref{subsec:exploration}).
The query instructs the LLM to infer the shelf scene's organization by jointly considering these attributes and to rank placement candidates anchored to existing shelf objects.
This allows the ranking to reflect scene-dependent attribute relevance without relying on a single attribute.
For the top-\(K\) semantic candidates, we convert the ordinal ranking into normalized exponential rank weights \(w_r\):
\begin{equation}
	w_r
	=
	\frac{
		e^{-(r-1)}
	}{
		\sum_{j=1}^{K}
		e^{-(j-1)}
	},
	\qquad r \in {1,\ldots,K}.
\end{equation}
The exponential decay prioritizes higher-ranked candidates while retaining lower-ranked alternatives when geometric constraints prevent the preferred placement or subsequent motion planning fails, as feasible grasp and placement configurations do not ensure a collision-free connecting trajectory.
To ground the semantic ranking in geometric placement poses, we sample \(100\) poses per shelf board on the coarse map with the target-object mesh~\cite{wingenderRobustPlaceabilityMetric2026}.
We associate each pose only with the candidates whose anchor objects lie on the same shelf board and within a planar distance of \(d_{\max}\!=\!\SI{25}{cm}\).
This restriction links semantic suitability to the local object arrangement.
Among eligible candidates, we select the association that maximizes the semantic score \(P_{\mathrm{sem}}\):
\begin{equation}
	P_{\mathrm{sem}}(p)
	=
	w_{r_i} 
	\left(
	1-\frac{d_i(p)}{d_{\max}}
	\right),
\end{equation}
where \(r_i\) is the ordinal rank of the hypothesis associated with pose \(p\), and \(d_i(p)\) is the distance between the placement position and its anchor object.
Ties are resolved in favor of the higher-ranked candidate.
The linear proximity term decreases from \(1\) at the anchor to \(0\) at the association threshold, favoring placements near anchors with high semantic ranks.
The resulting score can take values in the interval \([0,1]\).

To jointly decide how to pick and where to place, we then sort the grasp-placement pairs \((g,p)\) in descending order according to the score \(S\) defined as: 
\begin{equation}
	S(g,p)
	=
	G_{\mathrm{qual}}(g) \cdot P_{\mathrm{sem}}(p) \cdot \mathds{1}_{\mathrm{feas}}(g,p),
\end{equation}
where \(\mathds{1}_{\mathrm{feas}}(g,p)\) indicates whether the grasp-placement combination is collision-free and reachable, thereby excluding infeasible pairs.

Finally, the motion planner~\cite{sundaralingamCuRoboParallelizedCollisionFree2023} considers grasp-placement pairs in descending order of \(S(g,p)\) until it finds a collision-free trajectory for execution by the robot controller.

\section{Experimental Evaluation}
\label{sec:exp}
To demonstrate the performance of our approach in instance mapping accuracy and semantic placement coherence over single-view language grounding and geometry-only planning, we conducted experiments with a diverse set of packaged grocery and household objects in simulation and on a real robot.
We compare against a single-view LLM-based baseline~\cite{argenzianoEMPOWEREmbodiedMultirole2024} using the same segmenter, assessing label accuracy, geometric accuracy, and instance coverage.
We evaluate placement coherence through semantic fit against both a LLM baseline and a geometry-only pick-and-place baseline~\cite{wingenderRobustPlaceabilityMetric2026}, measuring execution success to assess whether greater placement coherence with the observed shelf organization is achieved while maintaining manipulation feasibility.
Across three scene sets, our approach improves all mapping metrics over the single-view baseline and achieves higher semantic fit than both baselines, with success rates comparable to geometry-only planning.
An additional attribute evaluation of the placement predictor highlights the benefits of LLM-based prediction over score-based prediction for finding semantically coherent placements.

\subsection{Baselines}
\label{subsec:baselines}
We use two different baseline approaches and one modified version of our approach for comparison.
The first baseline \textbf{(BL1)} predicts placements based on EMPOWER~\cite{argenzianoEMPOWEREmbodiedMultirole2024}.
For a fair comparison, we replace its original segmentation component with the same SAM\,3~\cite{carionSAM3Segment2026} object segmenter used in our approach.
In contrast to our method, BL1 operates on a single \mbox{RGB-D} observation of the shelf, as EMPOWER does not maintain a multi-view scene representation.
Moreover, EMPOWER derives object labels using part-of-speech~(POS) tagging, extracting nouns while discarding descriptive adjectives.
This can remove semantically relevant object information required for informed placement decisions.

The second baseline \textbf{(BL2)}, is a general pick-and-place pipeline based on UniP~\cite{wingenderRobustPlaceabilityMetric2026}, excluding its stability prediction component, as we manipulate packaged grocery and household items limiting reorientation to objects with planar side faces ensuring a stable support (see \secref{sec:problem_assumptions}).
BL2 uses the coarse map to sample geometrically feasible placement candidates but does not incorporate semantic information into the placement decision.

We further evaluate a variant of our approach (\textbf{Basic}), using an object descriptor that predicts labels from a basic prompt without detailed attributes as the full approach.

\subsection{Experimental Setup}
\label{subsec:setup}
For all experiments, we use the 6-DoF robot arm UR5e equipped with the Robotiq 2F-85 two-finger gripper as end-effector, which we both control with the Robot Operating System (ROS\,2)~\cite{macenskiRobotOperatingSystem2022a}.
To capture RGB-D observations, we use the Orbbec Gemini 336 camera (the Realsense L515 camera in simulation) mounted on the robot's wrist.
Furthermore, we rely on LangChain as LLM/VLM API with Pydantic for validation, NetworkX~\cite{hagbergExploringNetworkStructure2008} to model the scene graph, and Isaac Sim~\cite{gaoNVIDIAIsaacSim2026} for simulation experiments providing photo-realistic rendering.
As LLM/VLM models for our approach, we used OpenAI's GPT-5.6 Sol model and as LLM experts we used Mistral's Medium 3.5 model and Deepseek's V4.1-Flash model.
We executed all experiments on a desktop computer equipped with an AMD Ryzen 9 7900X3D CPU, \(128\)\,GB RAM, and an Nvidia RTX 4080 GPU.

\subsubsection*{\textbf{Scene Dataset}}
For the simulation experiments, we created a diverse dataset containing 60 packaged grocery and household objects divided into eight non-exclusive categories (baking and pantry, canned goods, condiments, cleaning, dairy, dry goods, personal care, and shelf stable drinks) and five different packaging types (box, bottle, can, carton, and tube) containing edible and non-edible products.

\begin{figure}[t]
    \vspace{0.25\baselineskip}
    \centering
    \includegraphics[width=0.75\linewidth]{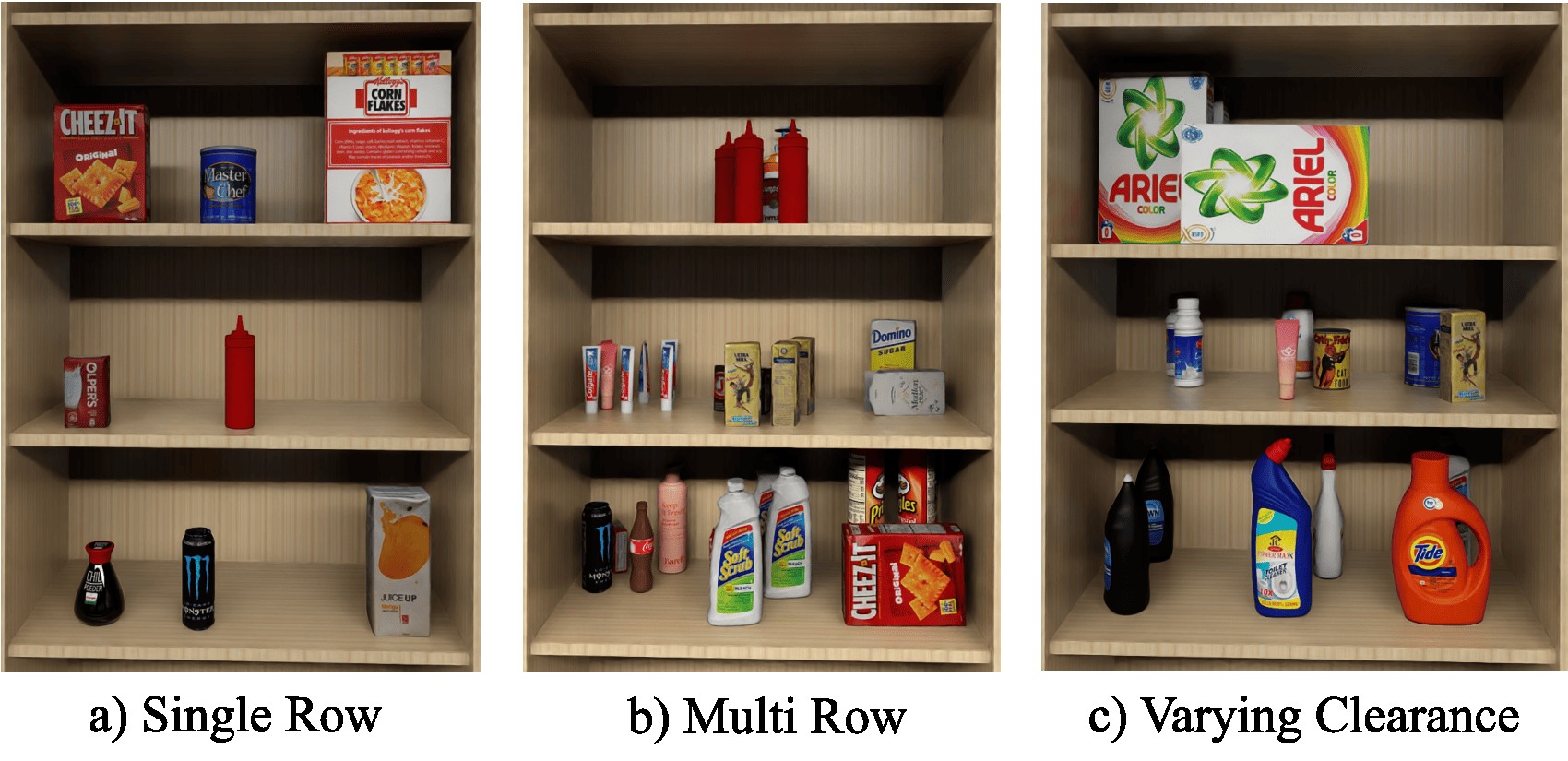}
    \vspace{-0.5\baselineskip}
    \caption{
    Simulation dataset with three procedurally generated scene configurations: (a) single-row scenes with objects arranged along the shelf width, (b) multi-row scenes with objects distributed across the shelf depth, introducing occlusions, and (c) varying-clearance scenes with additional variation in shelf-board heights and available clearance.
    }
    \label{fig:dataset}
\end{figure}
We procedurally generate three scene sets of 100 scenes each, referred to as single-row, multi-row, and varying-clearance scenes (see \figref{fig:dataset}).
In single-row scenes object are standing next to each other with a minimal random yaw perturbation of \SI{10}{\degree}, representing scenes comparable to other recent semantic placement works~\cite{argenzianoEMPOWEREmbodiedMultirole2024, wangAPRICOTActivePreference2024a}.
In multi-row scenes, we sample objects over the entire depth of each shelf board, resulting in objects standing behind each other, which leads to challenging object occlusions.
In the varying-clearance scenes, we also sample objects in multiple rows, but additionally vary the shelf board heights, limiting the available clearance.
We randomly draw objects from the complete dataset with sampling biased toward local groupings of objects from the same category to obtain realistic scenes.
Candidate poses are accepted only if they satisfy shelf bounds, height constraints, and collision-free placement.
In addition, we reserve free space for the target object to ensure its placability at multiple shelf locations.

\subsection{Metrics}
We evaluate our pipeline over 100 runs per scene set using the following metrics measuring execution success, semantic placement quality, and instance mapping accuracy.
For all metrics, we compared the paired per-run values using the two-sided Wilcoxon signed-rank test~\cite{wilcoxonIndividualComparisonsRanking1945} to check for statistical significance between our and all other approaches.

\subsubsection*{\textbf{Success Rate (SR)}}
We measure the percentage of successful runs.
A run is successful if all components (see \figref{fig:architecture}) can be executed and the robot places the object within \SI{6}{cm} of the planned placement pose.

\subsubsection*{\textbf{Semantic Fit (SF)}}
The metric measures the semantic suitability of a placement relative to nearby objects on the same shelf board.
To give closer neighbors more impact, we use Gaussian weighting of the planar Euclidean distance \(d\) including up to three closest neighbors~\(j\) using \mbox{\(\alpha_j\!=\!\exp[-d_j^2/(2\sigma^2)]\)} with \mbox{\(\sigma\!=\!0.25\)}.
Each component~\mbox{\(X\in\{C,P,E,O\}\)} is aggregated as \mbox{\(X\!=\!\sum_j\alpha_jX_j/\sum_j\alpha_j\)}.
The first content similarity component is defined as \(C_j\!=\!\max(0,\min(1,\cos(e_t,e_j)))\), where \(e_t\) and \(e_j\) are SBERT embeddings \cite{reimersSentenceBERTSentenceEmbeddings2019} of the target and neighboring object categories.
The edibility compatibility \(E_j\) is \(1\) for the same edible/non-edible group, \(0\) for different groups, and \(0.5\) if either group is unknown.
The packaging similarity \(P_j\) is~\(1\) if the target and neighboring objects have the same packaging type and \(0\) otherwise.
The orientation coherence \(O_j\) is \(1\) if their orientations (upright or lying) match and \(0\) otherwise.
We sum up the weighted components to the overall semantic fit score SF:
\begin{equation}
\label{eq:SF}
\mathrm{SF} =
w_c \cdot C + w_p \cdot P + w_e \cdot E + w_o \cdot O,
\end{equation}
where \(w_i\) are weights of each component.
We use two variants of this metric, where SF-EQ use an equal weight of \(0.25\) for all components and SF-LLM uses weights determined per scene by three individual LLM experts. 
Equal weights assess complementary aspects of shelf organization without imposing a predefined attribute hierarchy, while the LLM-based metric captures the latent shelf organization.
Each component can take values in the interval \([0,1]\), where higher values indicate a more semantically coherent placement.
\begin{table*}[t]
        \begingroup
        \scriptsize
        \setlength{\tabcolsep}{0.4pt}
        \centering
            \begin{tabularx}{\linewidth}{l*{12}{|Y}}
            \toprule
            \multicolumn{1}{c|}{\multirow{2}{*}{\textbf{Metric}}} & \multicolumn{4}{c|}{\textbf{Single Row Scenes}} & \multicolumn{4}{c|}{\textbf{Multi Row Scenes}} & \multicolumn{4}{c}{\textbf{Varying Clearance Scenes}}
            \\
            & \multicolumn{1}{c}{\textbf{GT-Labels}} & \multicolumn{1}{c}{\textbf{OURS}} & \multicolumn{1}{c}{\textbf{Basic}} & \multicolumn{1}{c|}{\textbf{BL1}}
            & \multicolumn{1}{c}{\textbf{GT-Labels}} & \multicolumn{1}{c}{\textbf{OURS}} & \multicolumn{1}{c}{\textbf{Basic}} & \multicolumn{1}{c|}{\textbf{BL1}}
            & \multicolumn{1}{c}{\textbf{GT-Labels}} & \multicolumn{1}{c}{\textbf{OURS}} & \multicolumn{1}{c}{\textbf{Basic}} & \multicolumn{1}{c}{\textbf{BL1}}
            \\
            \midrule
            a)\,F1-Label\,[\%]~\textuparrow
              & \(79 \pm 1.4\) & \(\mathbf{78} \pm 1.3\) & \(71 \pm 1.7^{\star}\) & \(36 \pm 2.1^{\star\star\star}\)
              & \(64 \pm 1.3\) & \(\mathbf{63} \pm 1.3\) & \(58 \pm 1.5^{\star}\) & \(24 \pm 1.6^{\star\star\star}\)
              & \(58 \pm 1.5\) & \(\mathbf{56} \pm 1.6\) & \(51 \pm 1.4^{\star}\) & \(24 \pm 1.5^{\star\star\star}\)
              \\
            b)\,BB-IoU\,[\%]~\textuparrow
              & \(63 \pm 0.8\) & \(\mathbf{63} \pm 0.9\) & \(62 \pm 1.2\) & \(21 \pm 0.1^{\star\star\star}\)
              & \(41 \pm 0.7\) & \(\mathbf{41} \pm 0.7\) & \(39 \pm 1.0\) & \(20 \pm 0.2^{\star\star\star}\)
              & \(40 \pm 1.1\) & \(\mathbf{40} \pm 1.0\) & \(39 \pm 1.2\) & \(21 \pm 0.1^{\star\star\star}\)
              \\
            c)\,Instances\,[\%]~\textuparrow
              & \(97 \pm 0.8\) & \(\mathbf{96} \pm 0.7\) & \(\mathbf{96} \pm 1.4\) & \(49 \pm 1.9^{\star\star\star}\)
              & \(80 \pm 1.0\) & \(\mathbf{80} \pm 0.8\) & \(\mathbf{80} \pm 1.6\) & \(29 \pm 1.2^{\star\star\star}\)
              & \(85 \pm 1.0\) & \(\mathbf{83} \pm 1.0\) & \(\mathbf{83} \pm 1.3\) & \(31 \pm 1.4^{\star\star\star}\)
              \\
            \bottomrule
            \end{tabularx}
        \endgroup
        \vspace{-0.5ex}
        
        \noindent
        \caption{%
        Instance mapping is evaluated over 100 runs per scene dataset using canonical-label F1, 3D bounding-box overlap (BB-IoU), and the number of mapped instances relative to the ground truth (Instances).
        GT-Labels provides an upper-bound reference by evaluating instance mapping with ground-truth object labels instead of predicted labels.
        OURS outperforms BL1~\cite{argenzianoEMPOWEREmbodiedMultirole2024} in all three metrics across all scene sets despite using the same SAM\,3 segmenter, supporting the combination of richer semantic descriptions and multi-view evidence.
        OURS also maps substantially more instances than BL1 across all scene sets, reaching \SI{80}{\percent} of the ground-truth instance count in multi-row scenes compared with \SI{29}{\percent} for BL1.
        BL1's noun-based label extraction likely contributes to the gap by discarding descriptive cues needed for object disambiguation.
        Compared with the attribute-free Basic variant, OURS improves F1 and matches or improves BB-IoU, supporting attribute-based instance association.
        \mbox{\(\star\) and \(\star\star\star\)} denote statistically significant results for our approach compared to all other approaches with \(p\)-value thresholds of \(0.05\) and \(0.001\) respectively.
        }
        \label{tab:quant_map_evaluation}
        \vspace{-0.75\baselineskip}
\end{table*}
\begin{table*}[t]
        \begingroup
        \scriptsize
        \setlength{\tabcolsep}{0.4pt}
        \centering
            \begin{tabularx}{\linewidth}{l*{9}{|Y}}
            \toprule
            \multicolumn{1}{c|}{\multirow{2}{*}{\textbf{Metric}}} & \multicolumn{3}{c|}{\textbf{Single Row Scenes}} & \multicolumn{3}{c|}{\textbf{Multi Row Scenes}} & \multicolumn{3}{c}{\textbf{Varying Clearance Scenes}}
            \\
            & \multicolumn{1}{c}{\textbf{OURS}} & \multicolumn{1}{c}{\textbf{BL1}} & \multicolumn{1}{c|}{\textbf{BL2}}
            & \multicolumn{1}{c}{\textbf{OURS}} & \multicolumn{1}{c}{\textbf{BL1}} & \multicolumn{1}{c|}{\textbf{BL2}}
            & \multicolumn{1}{c}{\textbf{OURS}} & \multicolumn{1}{c}{\textbf{BL1}} & \multicolumn{1}{c}{\textbf{BL2}}
            \\
            \midrule
            a)~SR~[\%]~\textuparrow
              & \(96 \pm 2.0\)  & \(20 \pm 4.0^{\star\star\star}\) & \(\mathbf{97} \pm 1.7\)
              & \(\mathbf{95} \pm 2.2\) & \(20 \pm 2.7^{\star\star\star}\) & \(\mathbf{95} \pm 2.1\)
              & \(\mathbf{81} \pm 3.9\) & \(5 \pm 2.2^{\star\star\star}\) & \(\mathbf{81} \pm 2.9\)
              \\
            b)~SF-EQ~[\%]~\textuparrow
              & \(\mathbf{60} \pm 1.5\) & \(35 \pm 2.6^{\star\star\star}\) & \(52 \pm 1.6^{\star\star\star}\)
              & \(\mathbf{63} \pm 1.5\) & \(34 \pm 2.7^{\star\star\star}\) & \(53 \pm 1.6^{\star\star\star}\)
              & \(\mathbf{61} \pm 2.2\) & \(37 \pm 3.0^{\star\star\star}\) & \(50 \pm 2.1^{\star\star\star}\)
              \\
            c)~SF-LLM~[\%]~\textuparrow 
              & \(\mathbf{58} \pm 1.7\) & \(32 \pm 2.6^{\star\star\star}\) & \(47 \pm 1.9^{\star\star\star}\)
              & \(\mathbf{58} \pm 1.6\) & \(29 \pm 2.5^{\star\star\star}\) & \(46 \pm 1.8^{\star\star\star}\)
              & \(\mathbf{58} \pm 2.2\) & \(33 \pm 2.8^{\star\star\star}\) & \(45 \pm 2.2^{\star\star\star}\)
              \\
            \bottomrule
            \end{tabularx}
        \endgroup
        \vspace{-0.5ex}
        
        \noindent
        \caption{%
        Results measure execution success (SR) and semantic fit with equal attribute weights (SF-EQ) and scene-dependent LLM weights (SF-LLM) based on 100 runs per scene set.
        OURS achieves the highest SF-EQ and SF-LLM across all scenes while retaining a success comparable to the geometry-only baseline BL2.
        BL1 achieves substantially lower success, SF-EQ, and SF-LLM, consistent with its limited instance coverage providing fewer object anchors for placement reasoning.
        \mbox{\(\star\star\star\)} denotes statistically significant results for our approach compared to BL1 and BL2 with \(p\)-value threshold of \(0.001\).
        }
        \label{tab:quant_pnp_evaluation}
        \vspace{0.5\baselineskip}
\end{table*}
\begin{table}[t]
        \begingroup
        \scriptsize
        \setlength{\tabcolsep}{0.2pt}
        \centering
            \begin{tabularx}{\linewidth}{l|Y*{7}{|Y}}
            \toprule
            \multicolumn{1}{c|}{\multirow{2}{*}{\textbf{Attribute}}} & 
            \multicolumn{2}{c|}{\textbf{Content}} & \multicolumn{2}{c|}{\textbf{Packaging}} & \multicolumn{2}{c|}{\textbf{Edible}} & \multicolumn{2}{c}{\textbf{Orientation}}
            \\
            &
            \multicolumn{1}{c}{\textbf{OURS}} & \multicolumn{1}{c|}{\textbf{Score}} &
            \multicolumn{1}{c}{\textbf{OURS}} & \multicolumn{1}{c|}{\textbf{Score}} &
            \multicolumn{1}{c}{\textbf{OURS}} & \multicolumn{1}{c|}{\textbf{Score}} &
            \multicolumn{1}{c}{\textbf{OURS}} & \multicolumn{1}{c}{\textbf{Score}}
            \\
            \midrule
            SF-LLM~[\%]~\textuparrow &
            \(\mathbf{72} \pm 2.2\) & \(68 \pm 2.4^{\star}\) &
            \(\mathbf{83} \pm 2.1\) & \(78 \pm 2.1^{\star}\) &
            \(\mathbf{70} \pm 2.0\) & \(66 \pm 2.2^{\star}\) &
            \(\mathbf{51} \pm 3.8\) & \(41 \pm 3.6^{\star\star}\)
            \\
            \bottomrule
            \end{tabularx}
        \endgroup
        \vspace{-0.5ex}
        
        \noindent
        \caption{%
        We determine attribute weights with three independent LLM experts using four sets of ten scenes, each dominated by one attribute while randomizing the others.
        Using these weights on 20 unseen scene sets, we compare our LLM-based predictor (OURS), against an equally weighted score-based predictor (Score).
        The results show that our predictor captures scene-specific attribute importance and produces more attribute-coherent placements without manual weight tuning.
        \mbox{\(\star\) and \(\star\star\)} denote statistically significant results for our approach compared to the score-based predictor with \(p\)-value thresholds of \(0.05\) and \(0.01\) respectively.
        }
        \label{tab:quant_score_evaluation}
\end{table}

\subsubsection*{\textbf{Instance Mapping (F1-Label/BB-IoU/Instances)}}
To evaluate the instance mapping, we report the F1 score of canonical object labels (F1-Label)
and the three-dimensional bounding box intersection over union (BB-IoU) compared to the ground truth.
For F1-Label, we consider the cosine similarity of SBERT embeddings between ground truth and predicted labels with a threshold of \SI{60}{\percent}.
We further report the amount of mapped instances (Instances) compared to the ground truth.

\subsection{Experimental Results}
\label{subsec:results}
We evaluate instance mapping (\textbf{OURS}) against the baseline (\textbf{BL1}) and the basic object descriptor (\textbf{Basic}), and placement quality against both the semantic-baseline BL1 and the geometry-only baseline (\textbf{BL2}).

\subsubsection*{\textbf{Quantitative Instance Mapping Results}}
\tabref{tab:quant_map_evaluation} shows that OURS outperforms BL1 in canonical-label F1, BB-IoU, and instance coverage across all scenes.
For multi-row scenes, F1 increases from \SI{24}{\percent} to \SI{63}{\percent} and the mapped instance count from \SI{29}{\percent} to \SI{80}{\percent} of the ground truth.
Since both methods use the same segmenter, the differences stem from mapping and association, indicating the benefit of multi-view observations and richer semantic descriptions.
Multiple viewpoints increase scene coverage by revealing objects that are occluded or only partially visible in a single view, while additional object attributes provide cues for cross-view disambiguation.
BL1, in contrast, uses a single observation and its POS-based label extraction removes descriptive adjectives, reducing both scene coverage and label specificity.

The comparison with Basic shows the effect of the semantic attributes on label association.
OURS obtains significantly higher F1 across all scenes, while achieving similar BB-IoU and map nearly the same number of instances. 
The attributes therefore improve cross-view association and semantic label consistency, additional to geometric reconstruction and instance coverage.
OURS also approaches GT-Labels in F1 while matching its BB-IoU, showing that the proposed semantic grounding recovers most of the association performance obtained with ground-truth labels.
\begin{figure*}[t]
    \centering
    \begingroup
    \newcommand{\rolloutPath}{figures/rollout}

    \newcommand{\rolloutDots}{%
        \ensuremath{{\cdot}\mkern-2mu{\cdot}\mkern-2mu{\cdot}}%
    }

    \begin{tikzpicture}[
        x=1.0\textwidth,
        y=1.0\textwidth,
        image/.style={
            inner sep=0pt,
            outer sep=0pt
        },
        heading/.style={
            inner sep=0pt,
            outer sep=0pt,
            anchor=south,
            font=\fontsize{6}{7}\selectfont\bfseries,
            text height=1.5ex,
            text depth=0pt
        }
    ]
        \node[heading] (targetHeader)
            at (0.070,0.004) {Target View};
        \node[heading] (shelfHeader)
            at (0.390,0.004) {Shelf Views};
        \node[heading] (representationHeader)
            at (0.722,0.004) {Representations};
        \node[heading] (executionHeader)
            at (0.909,0.004) {Execution};

        \foreach \view/\x in {
            0/0.070,
            1/0.230,
            2/0.380,
            6/0.550
        }{
            \node[image,anchor=north] at (\x,0) {
                \adjincludegraphics[
                    trim={0 0 0 {.25\height}},
                    clip,
                    width=0.119\textwidth
                ]{\rolloutPath/view_\view_seg.png}
            };

            \node[image,anchor=north] at (\x,-0.06) {
                \includegraphics[
                    trim=600bp 70bp 1400bp 140bp,
                    clip,
                    width=0.119\textwidth
                ]{\rolloutPath/view_\view.jpg}
            };
        }

        \node[image,font=\scriptsize] at (0.465,-0.034)
            {\rolloutDots};
        \node[image,font=\scriptsize] at (0.465,-0.145)
            {\rolloutDots};

        \node[image,anchor=north] at (0.722,0) {
            \includegraphics[
                width=0.136\textwidth,
                height=0.0408\textwidth,
                keepaspectratio
            ]{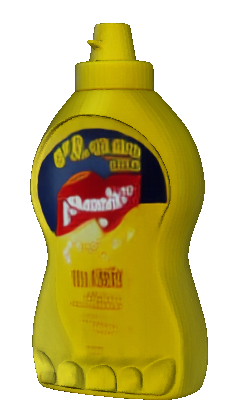}
        };

        \node[image,anchor=north] (instanceMap)
            at (0.722,-0.045) {
                \adjincludegraphics[
                    trim={0 {.11\height} 0 0},
                    clip,
                    width=0.136\textwidth,
                    height=0.1414145\textwidth,
                    keepaspectratio
                ]{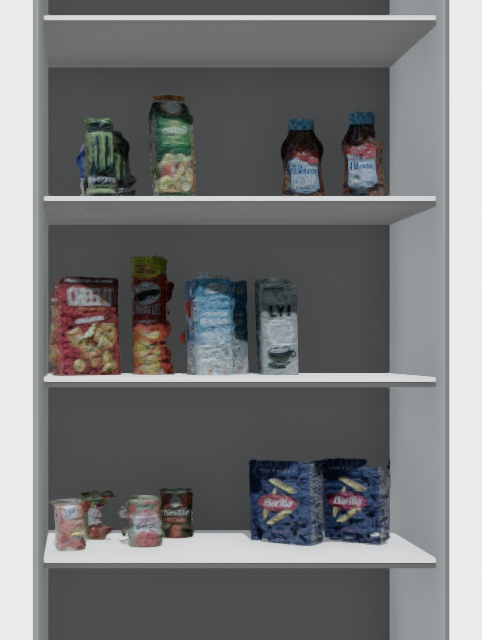}
            };

        \node[image,anchor=north] at (0.909,0) {
            \includegraphics[
                trim=600bp 70bp 1400bp 140bp,
                clip,
                width=0.153\textwidth,
                height=0.09068225\textwidth,
                keepaspectratio
            ]{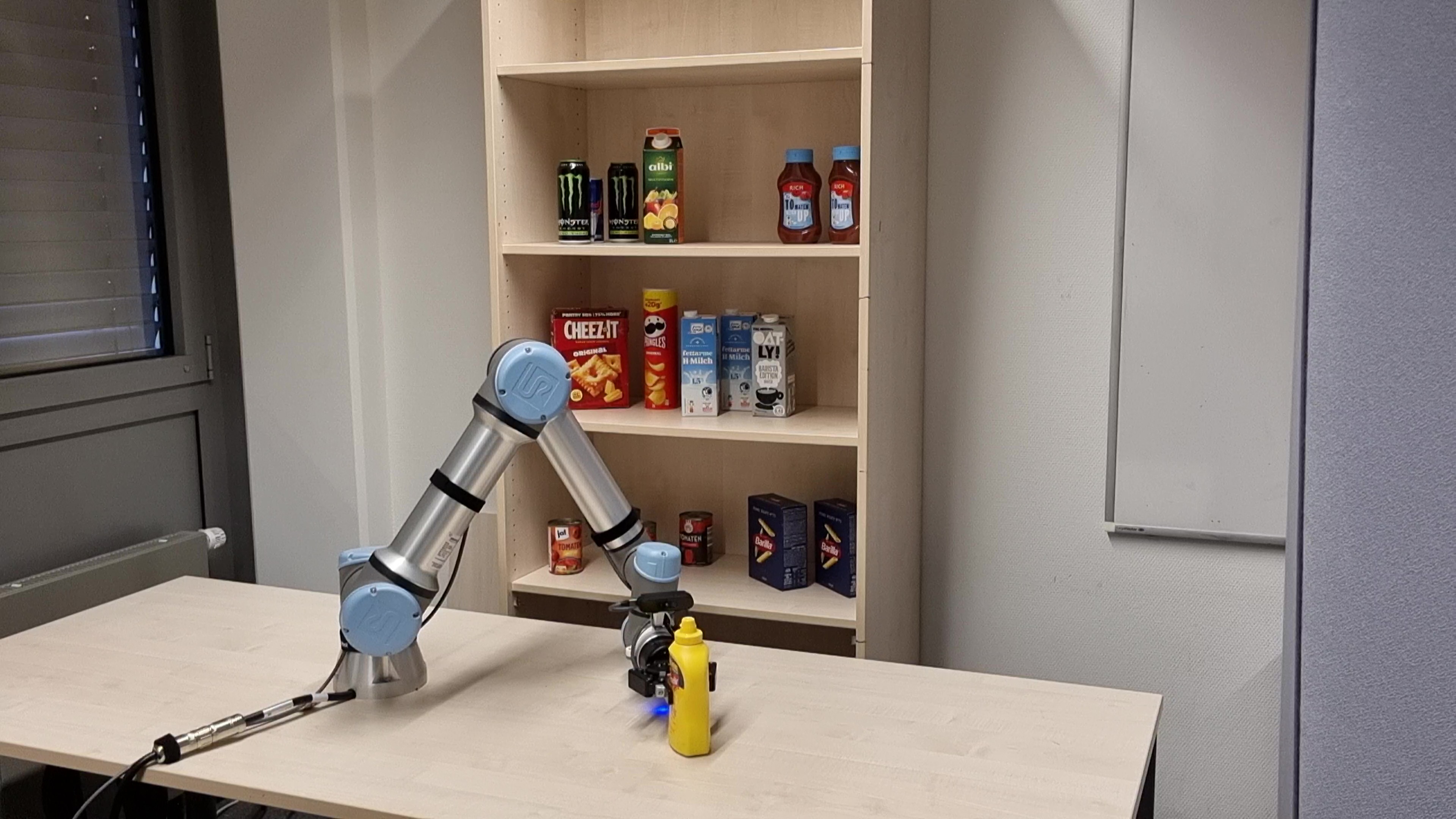}
        };

        \node[image,anchor=north] (placement)
            at (0.909,-0.096) {
                \includegraphics[
                    trim=600bp 70bp 1400bp 140bp,
                    clip,
                    width=0.153\textwidth,
                    height=0.09068225\textwidth,
                    keepaspectratio
                ]{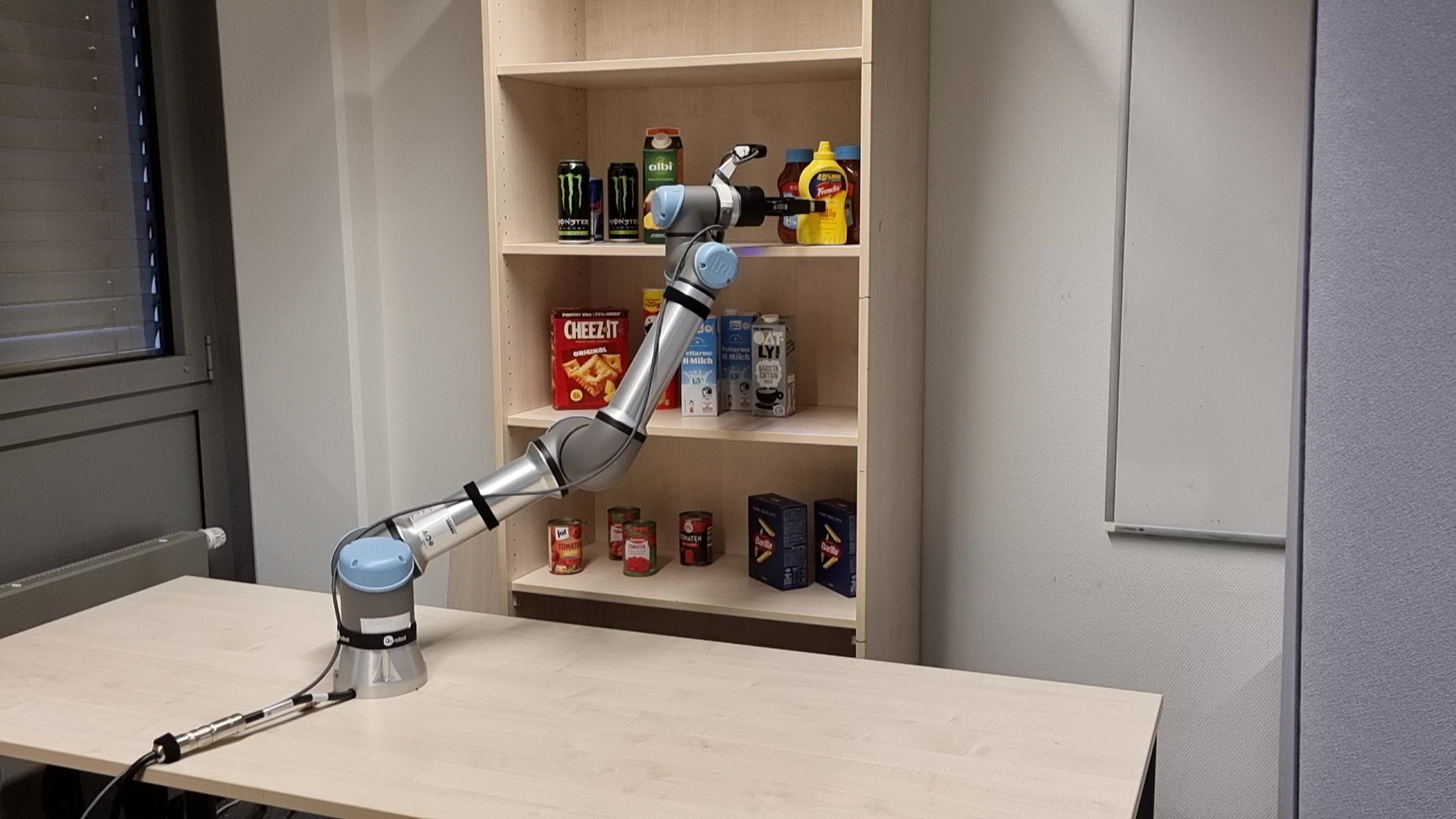}
            };

        \foreach \x in {0.150,0.640,0.806}{
            \draw[gray!60,line width=0.3pt]
                (\x,0 |- instanceMap.south) --
                (\x,0 |- targetHeader.north);
        }

        \path (0,0) (1,0);
    \end{tikzpicture}
    \endgroup
    \vspace{-1.25\baselineskip}
    \caption{
    We show an exemplary real-world rollout of the robot picking and placing a previously unseen object into a shelf.
    Target and shelf observations show segmentation results (top) and the corresponding robot configurations (bottom).
    The target object mesh supports geometric feasibility assessment, while the 3D semantic scene graph resulting from the depicted instance map encodes observed object attributes and spatial relations.
    Given this graph and the target attributes, the LLM ranks placements according to the inferred shelf organization.
    The executed top-ranked placement groups the target object with shelf objects of similar content and the same packaging type, consistent with the local organization of the shelf.
    }
    \label{fig:rollout}
\end{figure*}

\subsubsection*{\textbf{Quantitative Placement Results}}
\tabref{tab:quant_pnp_evaluation} shows that OURS achieves higher success, SF-EQ, and SF-LLM than BL1 across all scenes.
For multi-row scenes, SR increases from \SI{20}{\percent} to \SI{95}{\percent}, SF-EQ from \SI{34}{\percent} to \SI{63}{\percent}, and SF-LLM from \SI{29}{\percent} to \SI{58}{\percent}.
This agrees with the instance-mapping results, where the lower scene coverage of BL1 yields fewer object anchors and less context to infer the shelf organization, which limits the available placement relations.

Compared with the geometric-only baseline BL2, OURS achieves similar success but consistently higher SF-EQ and SF-LLM.
For multi-row scenes, both methods obtain an SR of \SI{95}{\percent}, while SF-EQ increases from \SI{53}{\percent} to \SI{63}{\percent} and SF-LLM from \SI{46}{\percent} to \SI{58}{\percent}.
BL2 samples geometrically feasible placements without semantic information, whereas OURS additionally reasons about the latent observed shelf organization.
The results therefore show that semantic multi-attribute reasoning improves placement consistency without reducing execution success.
The decrease in SR for both methods in varying-clearance scenes is consistent with the increased spatial limitations and object occlusions in the scenes, while OURS maintains similar SF-EQ and SF-LLM across all three scene sets.

\tabref{tab:quant_score_evaluation} compares our LLM-based placement predictor with an equally weighted score-based predictor (\textbf{Score}) across scene sets with different dominant attributes.
Score ranks sampled geometric placement poses using \eqref{eq:SF} with equal attribute weights, whereas OURS uses an LLM-based predictor.
This design provides a reference for scene-dependent attribute relevance while separating weight derivation on four sets of ten scenes from evaluation on 20 unseen sets.
OURS obtains mean SF-LLM scores of \SI{72}{\percent}, \SI{83}{\percent}, \SI{70}{\percent}, and \SI{51}{\percent} for content, packaging, edibility, and orientation, respectively, exceeding Score by \(4\), \(5\), \(4\), and \(10\) percentage points.
Equal weighting can favor agreement on several incidental attributes over the attribute that characterizes the shelf organization.
The observed gains are consistent with our predictor accounting for this scene-dependent priority without manual weight tuning.

\subsubsection*{\textbf{Qualitative Results}}
\figref{fig:rollout} shows an exemplary real-world rollout of our pipeline picking and placing a previously unseen object into a shelf.
The example demonstrates the integration of multi-view perception, semantic reasoning, and feasibility-aware planning beyond simulation.

\section{Conclusion}
\label{sec:conclusion}
We presented an open-world multi-attribute reasoning pipeline for placing previously unseen objects consistently with an existing shelf organization, without user-provided preferences or predefined object-receptacle mappings.
Our pipeline grounds object labels and attributes across views in a 3D semantic scene graph, which enables LLM-based semantic placement reasoning followed by geometric and kinematic feasibility checks.
Quantitative results show improved instance mapping and higher semantic fit over single-view language grounding and geometry-only planning.

\bibliographystyle{IEEEtran}
\bibliography{bibliography}
\end{document}